\documentclass[journal]{IEEEtran}

\IEEEoverridecommandlockouts                              

\usepackage{graphicx}
\usepackage{subfigure}
\usepackage{lipsum}
\usepackage{cite}
\usepackage{verbatim}
\usepackage{amsmath}
\usepackage{setspace}
\usepackage{fancyhdr}
\usepackage{hyperref}
\usepackage{color}
\usepackage{amsfonts}
\usepackage{multirow}
\usepackage[ruled]{algorithm2e} 
\usepackage{url}
\usepackage{svg}

\usepackage{bm}

\graphicspath{{./fig/}}
\DeclareGraphicsExtensions{.eps}

\title{Soft Robotic Mannequin: Design and Algorithm for Deformation Control}

\author{Yingjun Tian, Guoxin Fang,~\IEEEmembership{Member,~IEEE}, Justas S. Petrulis, 
Andrew Weightman, \\
and Charlie C.L. Wang$^\dag$,~\IEEEmembership{Senior Member,~IEEE}
\thanks{All authors are with the Department of Mechanical, Aerospace and Civil Engineering, The University of Manchester, United Kingdom.}
\thanks{$^\dag$Corresponding author: {\tt\small changling.wang@manchester.ac.uk}}
}

\usepackage{color}
\definecolor{amethyst}{rgb}{0.6, 0.4, 0.8}

\usepackage[normalem]{ulem}
\newcommand{\rev}[2]{{#2}}

\newcommand{\frev}[2]{\textcolor{red}{\sout{#1}}\textcolor{blue}{#2}}

\begin{document}
\maketitle

\bstctlcite{IEEEexample:BSTcontrol}

\begin{abstract}
This paper presents a novel soft robotic system for a deformable mannequin that can be employed to physically realize the 3D geometry of different human bodies. The soft membrane on a mannequin is deformed by inflating several curved chambers using pneumatic actuation. Controlling the freeform surface of a soft membrane by adjusting the pneumatic actuation in different chambers is challenging as the membrane's shape is commonly determined by the interaction between all chambers. Using vision feedback provided by a structured-light based 3D scanner, we developed an efficient algorithm to compute the optimized actuation of all chambers which could drive the soft membrane to deform into the best approximation of different target shapes. Our algorithm converges quickly by including pose estimation in the loop of optimization. The time-consuming step of evaluating derivatives on the deformable membrane is avoided by using the Broyden update when possible. The effectiveness of our soft robotic mannequin with controlled deformation has been verified in experiments. 
 
%
\end{abstract}


\section{Introduction}\label{secIntro}
\begin{figure}[t]
\centering
\includegraphics[width=\linewidth]{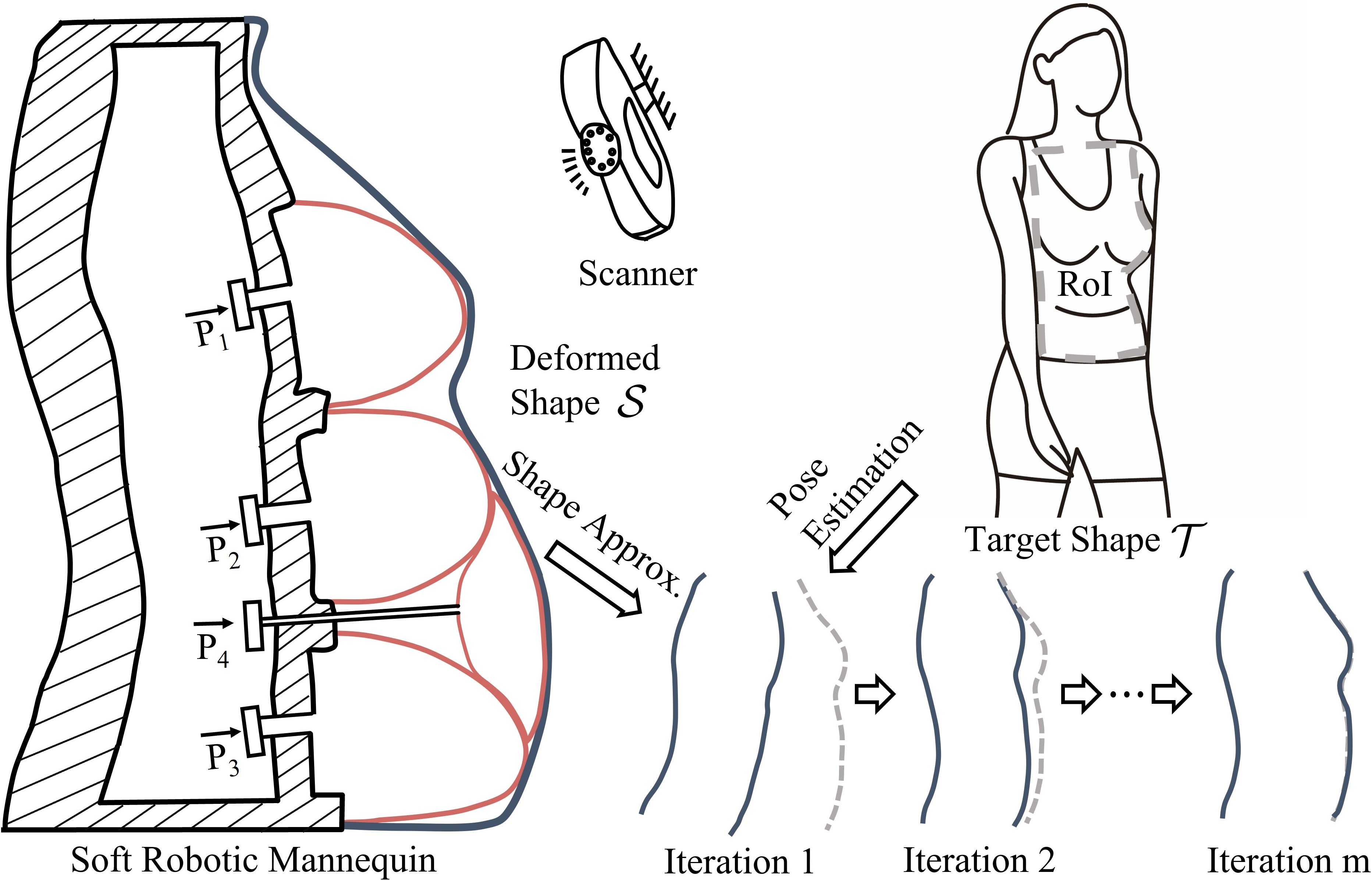}\\
\vspace{-5pt}
\caption{Given the \textit{region-of-interest} (RoI) on the body shape of a client as target $\mathcal{T}$, our soft robotic mannequin can be progressively deformed into a shape $\mathcal{S}$ that minimizes the difference between $\mathcal{T}$ and $\mathcal{S}$ by adjusting the pneumatic actuation $\{P_i\}$ of all chambers. A 3D scanner is employed in our system to provide vision feedback for deformation control. 
}\label{fig:WorkingPrinciple}
\end{figure}

\IEEEPARstart{T}{he} fabrication of tailor-made garments for a customer’s body shape \rev{}{currently} has to be matched most closely to a limited number of mannequins with prescribed geometries available in the workshop. Inevitably\frev{}{,} this leads to errors and prevents perfectly fitting garments. An ideal solution \rev{of}{to} this problem would be \rev{to realise }{}a deformable mannequin, the shape of which could be programmed according to the scanned body shapes of individual customers. Existing robotics mannequins such as Fit.me\cite{fitMe} and i.Dummy\cite{polyuDummy} are an improvement on traditional passive types but are limited by the large gaps formed by the non-continuous structures, which also result in imperfectly fitting garments. The actuation of these systems is provided by electrical motors which are sub-optimal solutions when the mannequin is used as the mold of a shaping iron with pressurized steam at a temperature up to $140^{\circ}\mathrm{C}$. In this paper, for the first time, we present a novel solution of a soft robotic deformable mannequin. The proposed soft robotic mannequin consists of a soft membrane which is deformed by inflating a number of pneumatically actuated curved chambers (see Fig.\ref{fig:WorkingPrinciple}). The actuation is optimized by computationally minimizing the shape difference between the current shape of the membrane (obtained from a \rev{structure-light}{structured-light} based 3D scanner) and a scanned 3D body shape of the client. 


\subsection{Related Work}
\subsubsection{Deformable Mannequin} In addition to the aforementioned commercially available robotic mannequins, the recently launched Euveka \cite{euveka_2021} has added an elastic ‘skin’ on top of the rigid components to fill the large gaps between non-continuous structures. Besides\rev{of}{} commercially available robotic mannequins, this is an active area of research. Abel and Kruusmaa~\cite{Abels2013ShapeControl,Abels2013FemaleModel} developed a solution akin to \cite{fitMe} which utilized less elastic materials but is still challenged by large gaps between sections; a similar design was employed by Xu et al.~\cite{Xu2018HardMannequin}. This same group of researchers later developed a deformable mannequin solution with its shape controlled by flexible belts~\cite{Li2019CAD}, where the surface shape between belts remained uncontrolled.

\subsubsection{Actuated Shape Morphing} Actuated shape morphing has been a topic studied more generally within robotics for many years. A widely adopted design is the robotic pinboard system as demonstrated in \cite{Relief2010TEI,inForm2013UIST,Je2021ElevateAW}. Recently, \rev{researches}{researchers} have developed arrays of pneumatic actuated chambers which have been used in visual and tactile displays~\cite{Stanley2015Haptics,Peele2019SORO,Okamura_CLoseLoopShapeControl}. Robertson et al.~\cite{Robertson2019ACM} constructed a system with an array of linear vacuum-powered soft pneumatic actuators used to transport different objects via shape changing. FESTO’s WaveHandling technology~\cite{festo2013} has similar functionality.  The surface shapes generated by these approaches are limited as it can only deform in a simple way. More complex deformations are required for a robotic mannequin which has to control the interaction among multiple chambers. We present a solution to this challenging problem within the paper. 


\subsubsection{Deformation Control \& Optimization} Siéfert et al.~\cite{Sifert2018NM} 
presented the shape morphing elastomer by embedding a network of specially designed airways so that it could realize controlled deformation to the target shape. 
Computational technology for inverse design has been developed by Skouras et al.~\cite{Skouras2014SIG} to realize deformation into desired target shapes. Along this thread of research, Ma et al.~\cite{Ma2017SIG} presented a method to compute 3D printed soft robots that can deform into a set of pre-defined shapes by the interaction among pneumatic actuated chambers. 
The challenge we seek to address in this paper is different. We aim to compute the optimized actuation for a soft mannequin \rev{utilising}{utilizing} vision feedback so that the mannequin deforms into a shape that provides the `best' approximation for an input target shape. Vision feedback has been employed to build closed-loop control for manipulating deformable objects~\cite{NavarroAlarcon2013ModelFreeVS, Han21_TMECH, Qi21_TMECH}, however, it has not been developed to control freeform deformation as \rev{what we proposed in this paper yet}{proposed}.


\subsection{Our Approach}\label{subsecOurApproach}
We have developed both the hardware and the software \rev{of}{for} a deformable mannequin as a soft robotic system. The working principle of our system is illustrated in Fig.\ref{fig:WorkingPrinciple}. Elastic chambers are fabricated on top of the core shape of a mannequin. By controlling the pressures of air (denoted by $\{ P_i \}$) pumped into these chambers, their shapes can be changed. A smooth mannequin body shape is achieved by attaching an elastic membrane on top of the other chambers. A structured-light based 3D scanner is employed to measure the current shape $\mathcal{S}$ of the mannequin as a point cloud. The difference between this 3D scanned mannequin shape and the desired target shape (input) is the error to be minimized. Details of our hardware design are presented in Section \ref{subsecHardware}. Without loss of generality, we can consider $\mathcal{S}$ as a function of the actuation parameters $\{ P_i \}$ -- i.e., $\mathcal{S}(P_1, \cdots, P_k)$ when there are $k$ chambers on the soft robotic mannequin.

Given a target shape $\mathcal{T}$ represented as a triangular mesh surface, we have developed an algorithm to adjust the values of $\{ P_i \}$ so that the shape difference between $\mathcal{T}$ and $\mathcal{S}$ is minimized. Specifically, the shape approximation function is defined as 
\begin{equation}\label{eqObjFunc}
    D(\mathcal{T},\mathcal{S}) = \sum_{j=1}^n \| \mathbf{p}_j - \mathbf{c}^{\mathcal{T}}_j \|^2 
    + \varpi \sum_{j=1}^n \| \mathbf{n}(\mathbf{p}_j) - \mathbf{n}(\mathbf{c}^{\mathcal{T}}_j) \|^2 
\end{equation}
where $\{\mathbf{p}_j\}$ are $n$ points sampled on the surface $\mathcal{S}$, $\mathbf{c}^{\mathcal{T}}_j$ gives the closest point of a query point $\mathbf{p}_j$ on the target shape $\mathcal{T}$, and $\mathbf{n}(\cdot)$ gives the surface normal at a point. $\varpi$ is a weight to balance the position error term and the normal error term. An optimization algorithm is developed to minimize the function $D(\cdot)$ therefore generating a `best' approximation of $\mathcal{T}$ by $\mathcal{S}$ (i.e., the deformable mannequin). The function value $D(\cdot)$ depends on both the actuation parameters $\{ P_i\}$ and the pose of target shape $\mathcal{T}$ denoted as a rigid transformation matrix $\mathbf{T}$. The optimization will be stuck at local optimum without considering the pose $\mathbf{T}$ (details can be found in Sec. \ref{subsecPoseEstimation}). We use the gradient-based method to minimize $D(\cdot)$ in our system (Sec. \ref{subsecGradientOptm}), and a method derived from the Broydan update is proposed to accelerate the process by evaluating fewer derivatives ${\partial D} / {\partial P_i}$ during iteration (Sec. \ref{subsecBroydanUpdate}). \rev{}{Broyden update is chosen because of its fast-convergence demonstrated in~\cite{NavarroAlarcon2013ModelFreeVS}.}

In summary, we make the following technical contributions:
\begin{itemize}
\item A novel design of a pneumatic actuated robotic system for a deformable mannequin as a soft robot. 
    
\item A closed-loop controller for deformation of the mannequin \rev{utilising}{utilizing} 3D vision-feedback.

\item An efficient and effective hybrid optimization algorithm including pose estimation for deformation control.
\end{itemize}
To the best of our knowledge, this is the first soft robotic system for mannequins with controlled deformation. The effectiveness of our approach has been verified in both simulation and physical experiments. 

\section{Pneumatic Actuated Deformable Mannequin}\label{secMannequin}

\begin{figure}[t] 
\centering
\includegraphics[width=\linewidth]{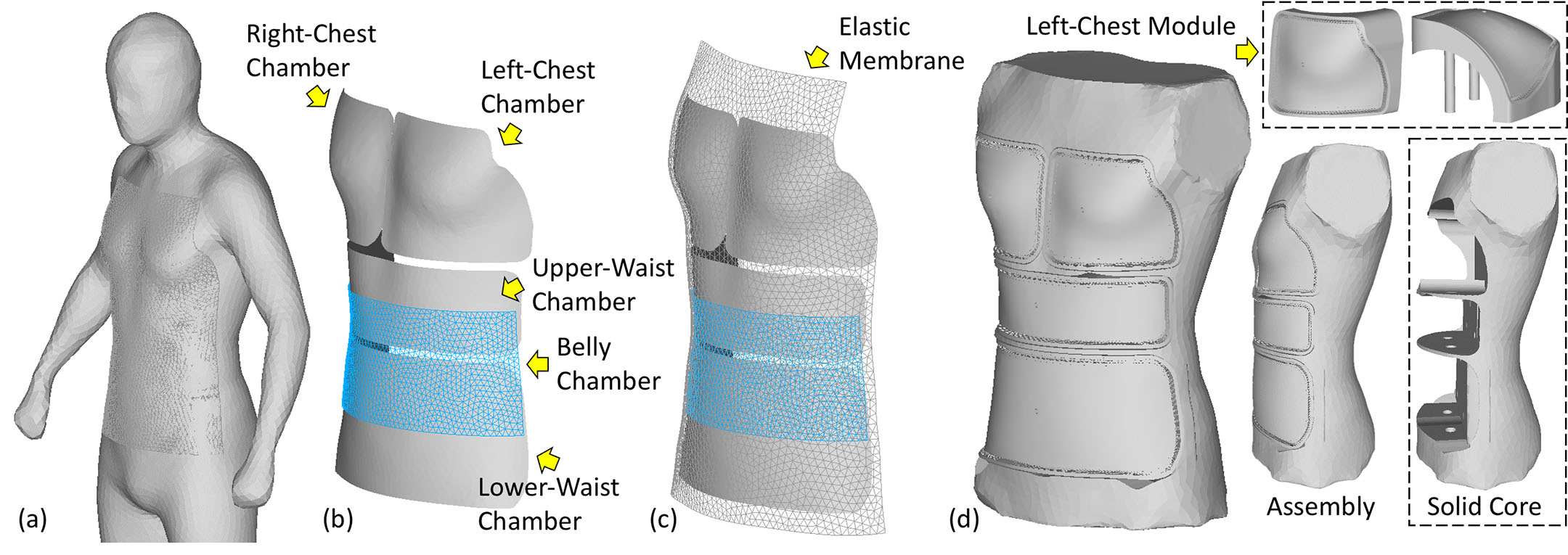}\\
\vspace{-5pt}
\caption{The chambers of our deformable mannequin are designed by considering the shape variation on human bodies. (a) We focus on the front upper-body for proof-of-concept research. (b) Four chambers are generated by segmentation together with an additional belly chamber attached on top of the lower-waist and the upper-waist chambers. (c) A piece of elastic membrane is employed as `skin' to generate a realistic shape commonly formed by the inflated chambers. (d) Every chamber except the belly chamber is designed as a solid module that can be first fabricated and then inserted into the solid core -- upper right shows the left-chest module as an example.
}\label{fig:chamberDivision}
\end{figure}

\subsection{Design and Fabrication}\label{subsecHardware}
For proof-of-concept research, we only focus on the front upper-body part of human \rev{body}{bodies} in this work. The designed soft robotic mannequin is expected to be deformed to approximate any human body shape at the same scale (i.e., similar height of the upper body). \rev{}{This design decision is based on the common practice of the garment industry that individuals are usually classified into different categories where height range is the most widely used parameter \cite{AldrichWinifred2015MPCf}.} 
For this purpose, we \rev{analyze}{analyzed} the shape space spanned by a 3D human body database (e.g., \cite{Chu2010,Reed2014}) and obtained an average body shape from models in the slimmest group (e.g., the model shown in Fig.\ref{fig:chamberDivision}(a)). Through shape analysis, we segmented the front of \rev{}{the} upper body into four regions of deformation (see the grey patches given in Fig.\ref{fig:chamberDivision}(b)), where each will become a chamber to be inflated. In order to realize a more realistic belly with fat, we \rev{design}{designed} an additional belly chamber on top of the lower / upper waist chambers (see the blue patch in Fig.\ref{fig:chamberDivision}(b)). However, this novel design of using an additional chamber also leads to challenges in control as the deformations of chambers interfere with each other. To generate a more realistic shape (commonly determined by the interaction between the chambers), we used an outer membrane as `skin' on top of all these chambers (see Fig.\ref{fig:chamberDivision}(c)). After segmentation, each region was converted into a solid block that could be \rev{insert}{inserted} into the solid core (as illustrated in Fig.\ref{fig:chamberDivision}(d)).
\rev{}{This modular design makes the whole mannequin easier to fabricate, repair and re-use.}

\begin{figure}[t] 
\centering
\includegraphics[width=1.0\linewidth]{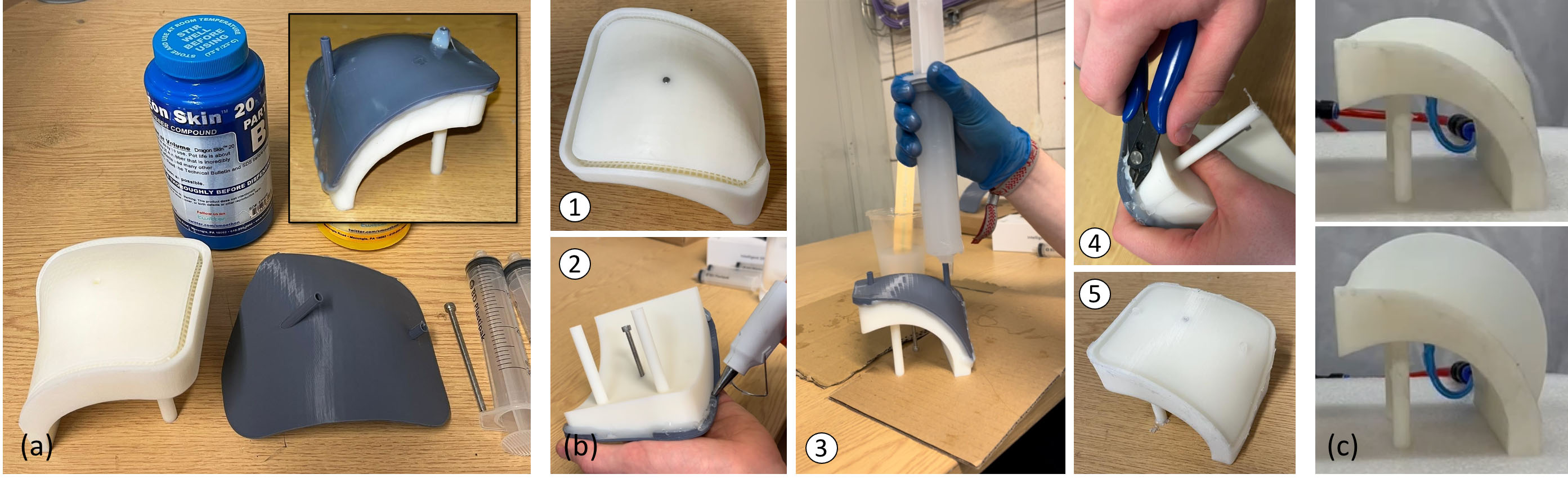}\\
\vspace{-8pt}
\caption{The chamber is fabricated by casting silicone rubber on top of the 3D printed PLA \rev{}{(}as \rev{}{the} lower part\rev{}{)}: (a) a 3D printed cover with inlets (gray) is designed to form the cavity of casting together with the 3D printed PLA solid (white), (b) steps for casting silicone rubber to form the upper part of \rev{}{the} chamber, and (c) the chamber can be easily inflated by pumping air through the hole of airway designed at the back of \rev{}{the} lower part.
}\label{fig:siliconeCasting}
\vspace{-5pt}
\end{figure}

\begin{figure}[t] 
\centering
\includegraphics[width=1.0\linewidth]{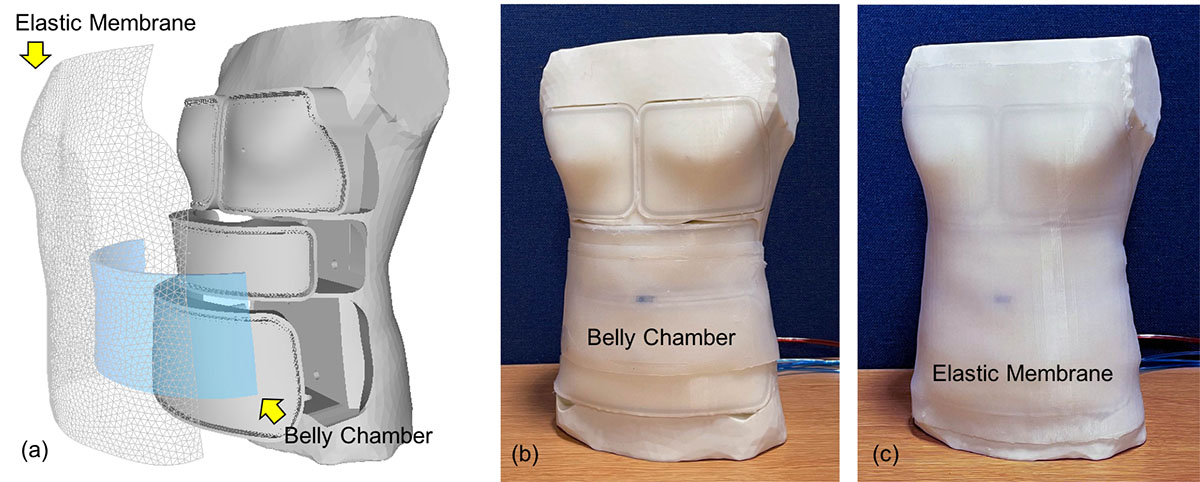}\\
\vspace{-8pt}
\caption{The fabrication of the mannequin is completed by assembling four different modules, the belly chamber and the elastic membrane (a). After attaching the belly chamber (b), the elastic membrane as `skin' is stitched on top of all chambers (c).
}\label{fig:mannequinAssembly}
\vspace{-10pt}
\end{figure}

The solid model of each module was converted into the lower part of the chamber, while the upper part of the chamber was fabricated by casting silicon around the lower part. Dragon Skin 20 silicone was adopted in our prototype. A cover with inlets was first designed to form a cavity for casting (see Fig.\ref{fig:siliconeCasting}). Note that silicone in general shows very weak chemical bonding stiffness compared to the 3D printed \textit{Polylactic Acid} (PLA) -- the material used to fabricate the lower part of the chamber. Micro-structures proposed in \cite{Rossing2020MicroBnd} were added at the boundary of each chamber to reinforce the air-tightness. For the region to form the cavity of a chamber, \rev{}{the} mould release agent was first sprayed before injecting the silicone liquid. After \rev{}{the} consolidation of the silicone rubber, an airtight chamber was formed by silicone (top) and PLA (bottom). An example of an inflated chest chamber made in this way is shown in Fig.\ref{fig:siliconeCasting}(c). After fabricating all solid chambers, the belly chamber was fabricated as a silicone `bag' and attached on top of the already assembled lower-waist and upper-waist modules (see Fig.\ref{fig:mannequinAssembly}(a) and (b)). The fabrication of the mannequin was finally completed by adding the elastic membrane as `skin' (see Fig.\ref{fig:mannequinAssembly}(c)), where the Ecoflex\texttrademark~00-30 silicone was selected because of its much smaller Young's modulus ($200\mathrm{Psi}$) than the Dragon Skin 20 silicone ($550\mathrm{Psi}$). Note that the left-chest and the right-chest chambers were connected by an airway so that they could always be under the same pressure to result in symmetric deformation.

\begin{figure}[t]
\centering
\includegraphics[width=1.0\linewidth]{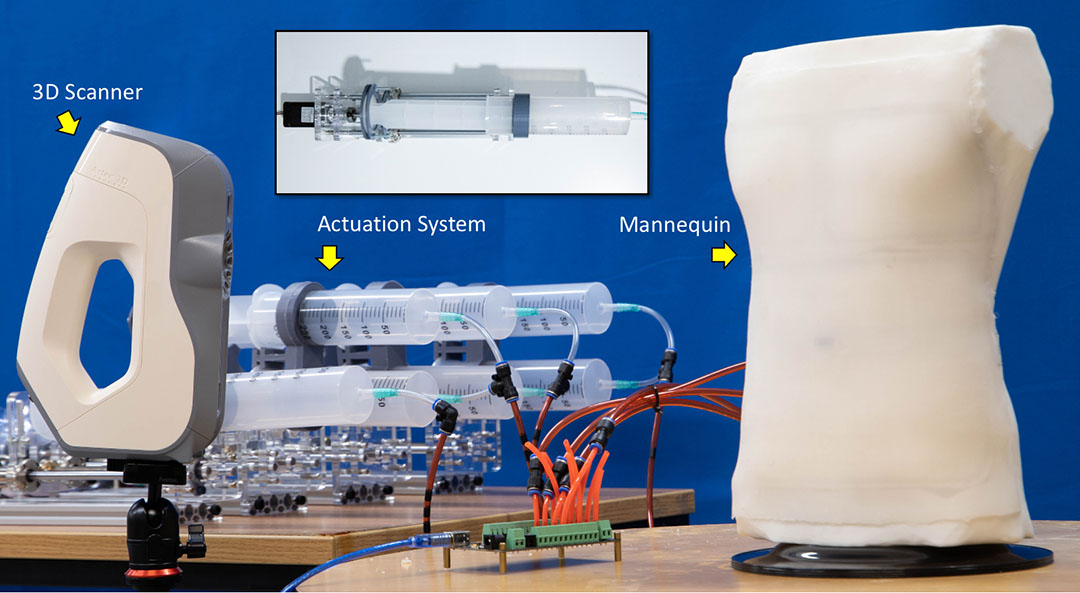}\\
\vspace{-5pt}
\caption{The hardware of our system with syringe-based pneumatic actuation (see also the top-view) and the structured-light based 3D scanner.
}\label{fig:hardwareSystem}
\vspace{-10pt}
\end{figure}
Our deformable mannequin is pneumatically actuated by a syringe system similar to that used in \cite{Fang2020TRO}, where the pressures of air pumped into the different chambers are controlled by the locations of plungers driven by DC motors (see the window-view in Fig.\ref{fig:hardwareSystem} for an illustration). Each chamber \rev{was}{is} equipped with a pressure sensor (Honeywell with pressure range: 60 psi), and the motors \rev{were}{are} driven by PID controllers to achieve the required air-pressures \rev{for different chambers}{at $30\mathrm{Hz}$}. A \rev{structure-light}{structured-light} based 3D scanner (i.e., Artec Eva Lite) is mounted in front of the mannequin. The shape of our mannequin in the region of interest can be obtained \rev{in real-time}{at the rate of $10\mathrm{Hz}$} as a mesh surface $\mathcal{S}$\rev{. This information}{, which} is used for deformation control.

\subsection{Challenges of Deformation Control}
It is challenging to develop a deformation control algorithm on this hardware system of a deformable mannequin. The major difficulties come from the following aspects.
\subsubsection{Under-actuation}
As a soft robotic system, the degree-of-freedom for the body shape is almost infinite but the dimension of our actuation space is limited (i.e., four independent chambers for the system designed above -- the chest, the upper-waist, the lower-waist and the belly chambers). The deformation control algorithm is developed based on the mannequin's current shape in an optimization framework. 

\subsubsection{Nonlinearity} The final shape of a mannequin $\mathcal{S}$ is commonly determined by the interaction between the chambers. The relationship between the shape of the mannequin $\mathcal{S}$ and the actuation parameter $\{P_i\}$ is highly nonlinear. Prior knowledge \rev{for}{of} the mechanical property of the chambers and the membrane is not available. Therefore, it is hard to formulate model-based kinematics. Moreover, the shape prediction model developed from physical simulation is far from the real results on physical mannequins because of the imprecisely controlled manufacturing process.

\subsubsection{Time-consuming gradient evaluation} When the deformation control is formulated in a framework of optimization (i.e., minimizing the objective function defined in Eq.(\ref{eqObjFunc})), the gradients need to be evaluated for the optimizer. The numerical difference \rev{}{evaluated} by running \rev{the evaluation on }{}the hardware system is very time-consuming.
\begin{equation}\label{eqGradientComponentf}
{\partial D}/{\partial  P_i} = (D(...,P_{i}+\Delta ,...)-D(...,P_{i},...)) / \Delta
\end{equation}
In short, we need to physically drive the actuation system $k$ times for a system with $k$ chambers which on average takes 12.5 sec. per actuation on our hardware.
\rev{}{Note that this is actually the control speed of our system as a whole, which is much slower than the inner pressure control loop and the camera system as the elastic chambers take a \rev{relative}{relatively} longer time to reach a stable shape -- i.e., the material hysteresis.}

We developed an algorithm using a hybrid solver for optimization, which requires less time for actuation on the physical system and shows similar if not fewer steps in iteration. 
\section{Actuation Optimization with Vision Feedback}\label{secOptimization}
\subsection{Gradient-Based Optimization}\label{subsecGradientOptm}
The primary scheme of our algorithm for deformation control was designed to achieve a target shape $\mathcal{T}$ based on the gradient descent equipped with a 1D search. The gradient of the shape approximation function ($D(\mathcal{T},\mathcal{S})$ as defined in Eq.(\ref{eqObjFunc})) with reference to the actuation (i.e., the controlled pressures $\mathbf{a}=\{P_i \}$ for different chambers) can be obtained by numerical difference as Eq.(\ref{eqGradientComponentf}) above. 

\begin{figure}[t]
\centering
\includegraphics[width=1.0\linewidth]{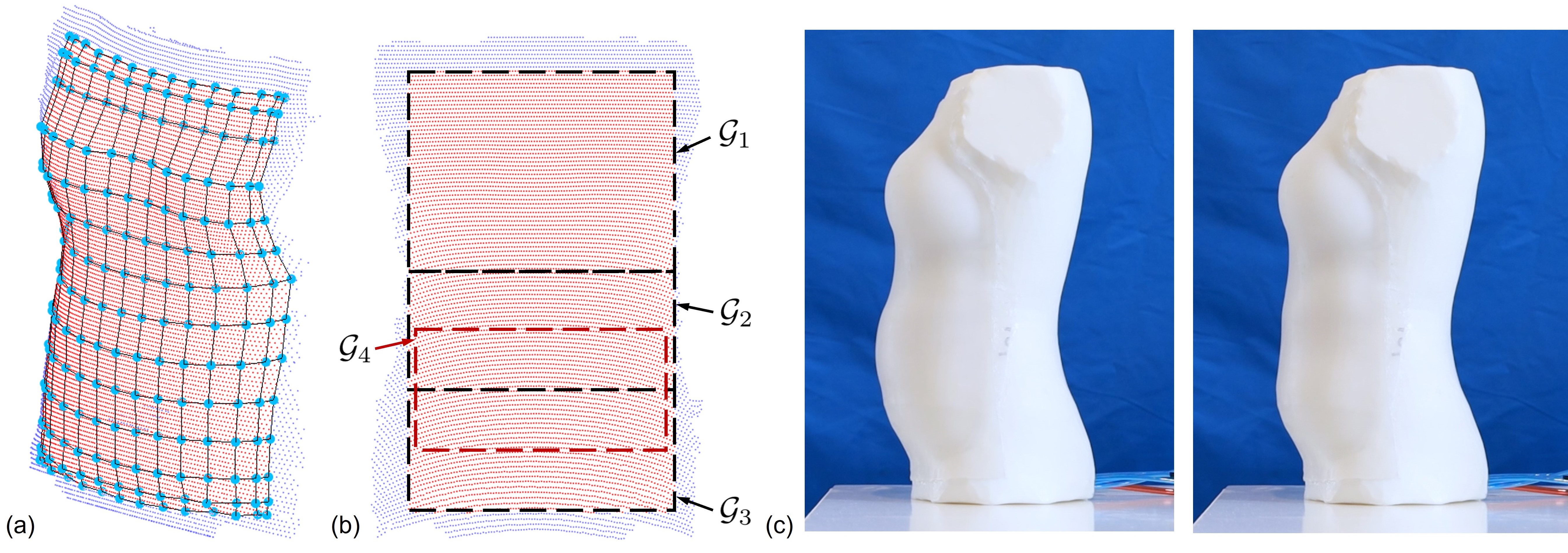}\\
\vspace{-5pt}
\caption{Region of interest (with points displayed in red) is parameterized by a B-spline surface (a) and also decomposed into four sub-regions (b), where each sub-region is mainly influenced by a corresponding chamber and can be predefined in the parametric domain. Without the additional belly chamber in the region $\mathcal{G}_4$, the shape in the belly region can hardly bulge -- see (c) for the different shapes generated with (left) vs. without (right) the belly chamber.}\label{fig:ParaDomainDecomp}
\end{figure}

The function value $D(\mathcal{T},\mathcal{S})$ depends on the shape of the mannequin captured by the 3D geometry scanner as a triangular mesh $\mathcal{M}$, which may embed noise. In order to reduce the influence of such noisy output from the scanner, we first fitted a cubic B-spline surface patch $\Tilde{\mathcal{S}}$ with $14 \times 14$ control points by using all vertices on $\mathcal{M}$ (see Fig.\ref{fig:ParaDomainDecomp}(a) for an example). Then, $n$ sample points predefined in the parametric domain of $\Tilde{\mathcal{S}}$ were projected back onto $\mathcal{M}$ -- i.e., $13 \times 25$ samples were uniformly sampled in the $uv$-domain of $\mathcal{M}$. These points $\{ \mathbf{p}_j\} \; (j=1,\cdots,n)$ were employed as the representative points of $\mathcal{S}$ for the function evaluation.

The vector of actuation $\mathbf{a}$ for realizing the target shape $\mathcal{T}$ on the physical mannequin can be determined by the following steps starting from an initial guess $\mathbf{a}_0$, where $D(\mathcal{T},\mathcal{S})$ is a function of $\mathbf{a}$ as $D(\mathcal{T},\mathcal{S}(\mathbf{a}))$. 
\begin{enumerate}
\item Computing the gradient vector $\mathbf{g}=\nabla D |_{\mathbf{a}=\mathbf{a}_i}$.

\item Determining a maximally allowed magnitude $\tau$ of $\mathbf{g}$ so that $(\mathbf{a}_i - \tau \mathbf{g})$ can be realized on all chambers of the mannequin.

\item Using the 1D-search strategy of shrinkage and expansion (ref.~\cite{Fang2020TRO}) to determine an optimal magnitude within the maximally allowed range as
\begin{equation}\label{eq1Dsearch}
\tau_{opt} = \arg \min_{\tau} D(\mathcal{T},\mathcal{S}(\mathbf{a}_i - \tau \mathbf{g})).
\end{equation}

\item Let $\mathbf{a}_{i+1}=\mathbf{a}_i - \tau_{opt} \mathbf{g}$ and go back to Step 1.
\end{enumerate}
We terminate the iteration by a hybrid condition when
\begin{equation}\label{eqTerminalCond}
{|D(\mathcal{T},\mathcal{S}(\mathbf{a}_{i+1}))-D(\mathcal{T},\mathcal{S}(\mathbf{a}_i))|}/{D(\mathcal{T},\mathcal{S}(\mathbf{a}_{i+1}))} \leq \tau
\end{equation}
with $\tau=1\%$ or it has reached the maximal steps as $i_{\max}=25$. 


\subsection{Pose Estimation}\label{subsecPoseEstimation}
\begin{figure}[t]
\centering
\includegraphics[width=\linewidth]{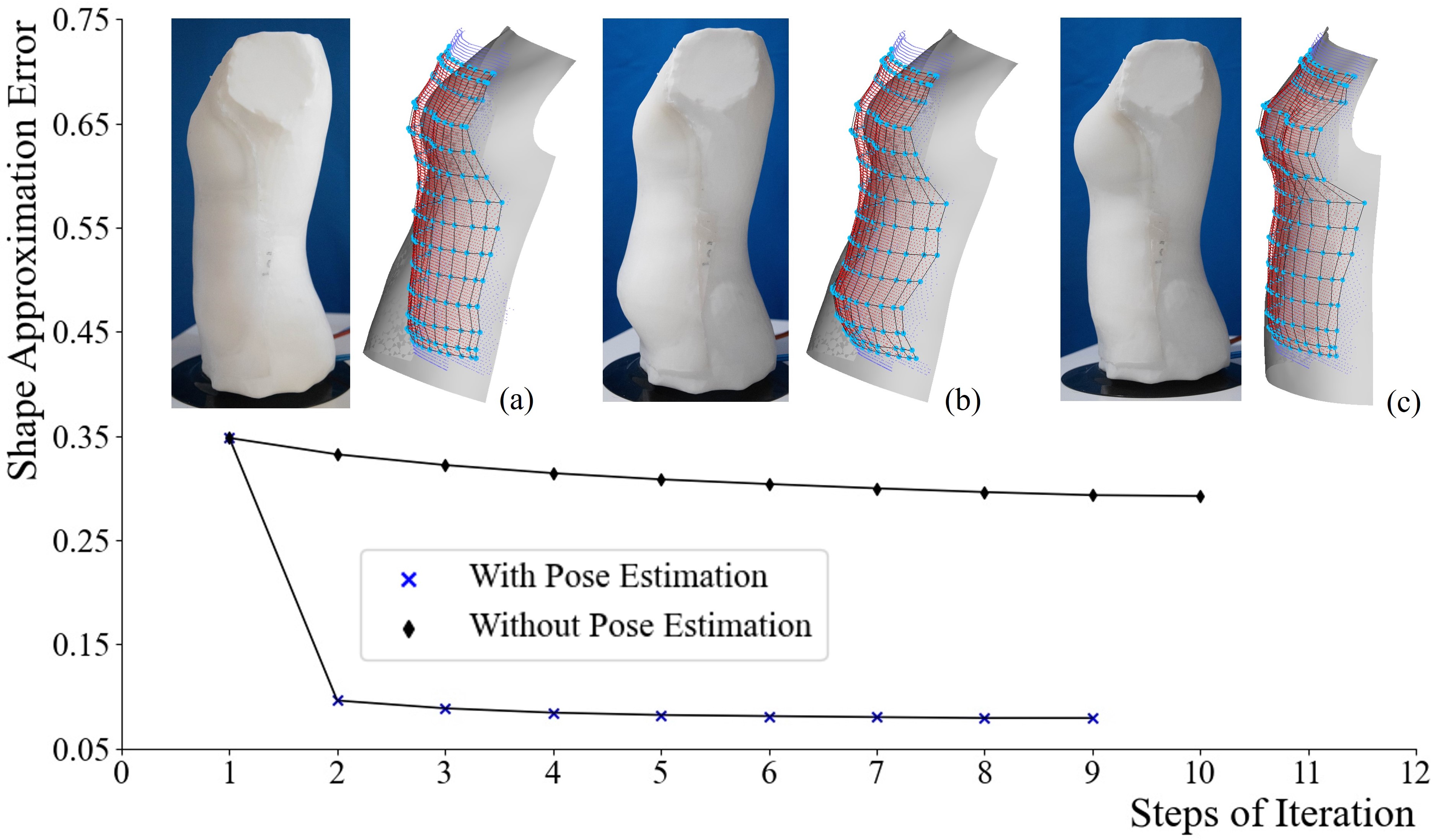}\\
\vspace{-5pt}
\caption{Pose estimation is a very important step for the deformation control of a soft robotic mannequin, where the target model is displayed as a grey mesh surface and the deformed shape of a mannequin is displayed as a point cloud together with the B-Spline surface. (a) The iteration starts from a misaligned pose between the target model and the mannequin. (b) The gradient-descent based optimization is stuck at the local optimum. (c) The converged result by incorporating the ICP-based pose estimation in the iteration.}\label{fig:MisalignmentWithWithoutICP}
\end{figure}

The gradient-descent based primary scheme did not consider the pose difference between the target shape $\mathcal{T}$ and the current shape $\mathcal{S}$ of the mannequin. The computation of deformation control based on optimization may be stuck at a local optimum and hard to converge when two models are misaligned in poses. For example, we made an extreme case of misalignment as shown in Fig.\ref{fig:MisalignmentWithWithoutICP}(a). Directly applying the gradient-descent optimizer will result in a strange shape (Fig.\ref{fig:MisalignmentWithWithoutICP}(b)).

To overcome the difficulty mentioned above, an \textit{iterative closest point} (ICP) based pose estimation step was incorporated in the computation of deformation control. The shape approximation function $D(\cdot)$ defined in Eq.(\ref{eqObjFunc}) was further considered as a function of both the actuation $\mathbf{a}$ and the pose of target shape $\mathbf{T}=(\mathbf{R},\mathbf{t})$ -- that is $D(\mathcal{T}(\mathbf{R},\mathbf{t}),\mathcal{S}(\mathbf{a}))$. However, it is impractical to optimize $D(\cdot)$ by changing all these variables together, which leads to a function with very high non-linearity. Instead, we adopted the strategy of bi-directional optimization to update the values of $(\mathbf{R},\mathbf{t})$ and $\mathbf{a}$ in different steps.

Specifically, the optimal pose $(\mathbf{R},\mathbf{t})$ was determined by considering $\mathcal{S}$ as static and minimizing the following function.
\begin{equation}\label{eqPoseEstimation}
D(\mathcal{T}(\mathbf{R},\mathbf{t}),\mathcal{S}) = \sum_{j=1,\cdots,n} \left\| \mathbf{p}_j - \left(\mathbf{R} \mathbf{c}^{\mathcal{T}^*}_j + \mathbf{t} \right) \right\|^2
\end{equation}
where $\mathbf{R}$ and $\mathbf{t}$ were updated progressively until a trivial update could be applied, and the closest point $\mathbf{c}^{\mathcal{T}^*}_j$ was searched on the target model $\mathcal{T}$ with pose updated. For the sake of simplicity in implementation, we neglect the normal term of Eq.(\ref{eqObjFunc}) here. This is the ICP algorithm widely used in rigid registration \cite{Rusinkiewicz2001}. We applied the ICP-based pose estimation to update $\mathbf{R}$ and $\mathbf{t}$ after every step of gradient-descent based optimization and found the computation converged very fast (e.g., the example shown in Fig.\ref{fig:MisalignmentWithWithoutICP}(c)).

\subsection{Acceleration by Broydan Update}\label{subsecBroydanUpdate}
Evaluating the gradient of objective function $D(\mathcal{T},\mathcal{S}(\mathbf{a}))$ with reference to the actuation vector $\mathbf{a}$ on the hardware setup is time-consuming. We reformulated the objective function defined in Eq.(\ref{eqObjFunc}) so that the strategy of Broydan update~\cite{Broyden1965} could be applied to reduce the need to \rev{evaluation}{evaluate} the gradient on the hardware when possible. An efficient algorithm was finally developed as a hybrid of gradient-descent and Broydan update integrated with ICP-based pose estimation. 

To enable the computation for Broydan update, we first decomposed the objective function $D(\cdot)$ into a function that returned a $k$-dimensional vector $\mathbf{d}$ as 
\begin{equation}
    \mathbf{d}(\mathcal{T},\mathcal{S})=[D_1 \; D_2 \; \cdots \; D_k]^T
\end{equation}
with $D(\mathcal{T},\mathcal{S})=\sum_{l=1}^k D_l$. Specifically, the region of interest was decomposed into four subregions: $\mathcal{G}_1$ for chest, $\mathcal{G}_2$ for upper-waist, $\mathcal{G}_3$ for lower-waist, and $\mathcal{G}_4$ for belly (see Fig.\ref{fig:ParaDomainDecomp}(b) for an illustration). The function $D_l$ is defined as
\begin{align}
    D_l(\mathcal{T},\mathcal{S}) = & \sum_{\mathbf{p}_j \in \mathcal{G}_l} w_j \| \mathbf{p}_j - \mathbf{c}^{\mathcal{T}}_j \|^2 \nonumber\\
    &+ \varpi \sum_{\mathbf{p}_j \in \mathcal{G}_l} w_j \| \mathbf{n}(\mathbf{p}_j) - \mathbf{n}(\mathbf{c}^{\mathcal{T}}_j) \|^2
\end{align}
by only using points belonging to the sub-region $\mathcal{G}_l$. For a point $\mathbf{p}_j$ that only belongs to a single sub-region, we use $w_j=1$. The sub-region of belly $\mathcal{G}_4$ overlaps with $\mathcal{G}_2$ and $\mathcal{G}_3$. Therefore, the weight $w_j=\frac{1}{2}$ is employed for point $\mathbf{p}_j \in \mathcal{G}_a \cap \mathcal{G}_b$ ($a \neq b$) to ensure $D(\mathcal{T},\mathcal{S})=\sum_{l=1}^k D_l$. 

The Jacobian matrix can then be obtained as
\begingroup
\begin{equation}\label{eqJacobianDiff}
\begin{aligned}
    \mathbf{J}  = \frac{\partial \mathbf{d}}{\partial  \mathbf{a}}=
    \Large
\left[
 \begin{matrix}
   \frac{\partial D_1}{\partial  P_1} & \frac{\partial D_1}{\partial  P_2}  &...\\
   \vdots&\ddots&\\
   \frac{\partial D_k}{\partial  P_1} &  & \frac{\partial D_k}{\partial  P_k}
  \end{matrix}
  \right].
\end{aligned}
\end{equation}
\endgroup
Starting from a known Jacobian $\mathbf{J}$, the following steps are applied to change the actuation vector $\mathbf{a}$ by the rules of a Broyden update.
\begin{equation}\label{eqBroydenIteration}
\mathbf{a}_{i+1} = \mathbf{a}_i - \lambda \mathbf{J}_i^{-1} \mathbf{d}_i
\end{equation}
where $\Delta \mathbf{a}_{i} = \mathbf{a}_{i} - \mathbf{a}_{i-1}$ and $\Delta \mathbf{d}_{i} = \mathbf{d}_{i} - \mathbf{d}_{i-1}$. Here $\lambda$ is a coefficient used to adjust the ratio of step size~\cite{NavarroAlarcon2013ModelFreeVS}. We chose $\lambda=0.1$ according to experimental tests. The inverse of the Jacobian is computed by the \textit{singular value decomposition} (SVD) in the first step, and it can be updated by using the following rule in the later steps.
\begin{equation}\label{eqInverseJacobianUpdate}
\mathbf{J}_{i}^{-1} = \mathbf{J}_{i-1}^{-1} + \frac{\Delta \mathbf{a}_{i} - \mathbf{J}_{i-1}^{-1} \Delta \mathbf{d}_{i} }{\mathbf{a}_{i}^T \mathbf{J}_{i-1}^{-1} \Delta \mathbf{d}_{i}} \Delta \mathbf{a}_{i}^T \mathbf{J}_{i-1}^{-1}
\end{equation}

\begin{figure}[t] 
\centering
\includegraphics[width=\linewidth]{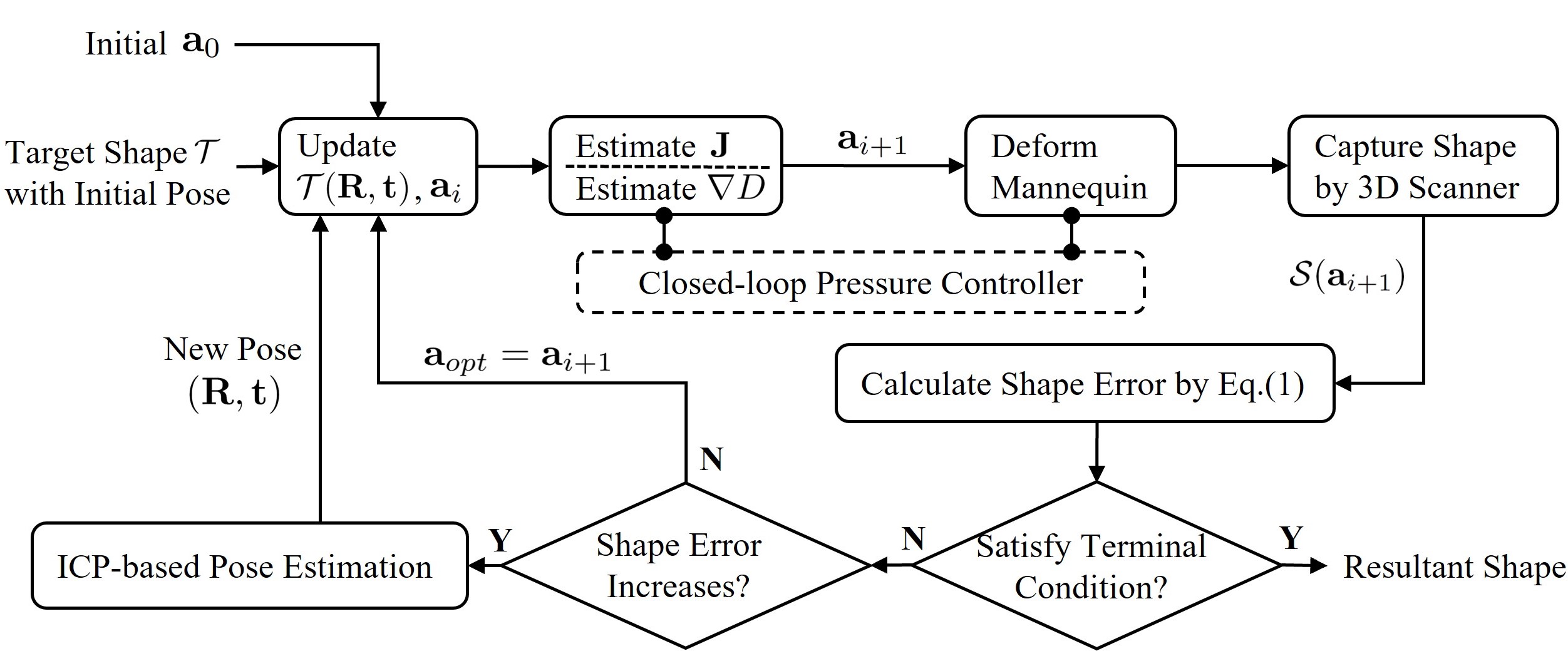}\\
\vspace{-5pt}
\caption{\rev{}{The block diagram of our deformation control algorithm.}}\label{fig:CtrlAlgorithm} 
\end{figure}

Applying the above rules to update the value of the actuation vector $\mathbf{a}$ did not always result in a configuration with a smaller shape approximation error (i.e., the value of $D(\mathcal{T},\mathcal{S})$). To resolve this problem, a hybrid scheme was developed to switch back to the gradient-descent when $D(\mathcal{T},\mathcal{S}(\mathbf{a}_{i+1})) > D(\mathcal{T},\mathcal{S}(\mathbf{a}_i))$. 
According to our experimental tests, the Broyden update can always further reduce the value of $D(\cdot)$ after applying 1-2 steps of gradient-descent. The pseudo-code of our hybrid algorithm is given in \textbf{Algorithm} \textit{HybridDefSolver}\rev{.}{, and the diagram of our control algorithm is shown in Fig.\ref{fig:CtrlAlgorithm}.}

\begin{algorithm}[t]\label{alg:hybridSolver}
\caption{HybridDefSolver}

\LinesNumbered

\KwIn{The target shape $\mathcal{T}$ and the initial actuation $\mathbf{a}_0$.}

\KwOut{The optimized actuation $\mathbf{a}_{opt}$, the pose of $\mathcal{T}$ as $(\mathbf{R},\mathbf{t})$ and the resultant shape $\mathcal{S}$.} 

Set $i = 0$ and apply the initial actuation vector $\mathbf{a_0}$;

\While{$i<i_{max}$}{

ICP-based pose estimation to get $(\mathbf{R},\mathbf{t})$ for $\mathcal{T}$;

\tcc{\small Gradient-descent solver (GDS)}

Compute the gradient vector $\mathbf{g}=\nabla D |_{\mathbf{a}=\mathbf{a}_i}$;

Apply the 1D-search to find the optimal update $\mathbf{a}_{i+1}=\mathbf{a}_i - \tau_{opt} \mathbf{g}$ with $\tau_{opt}$ determined by Eq.(\ref{eq1Dsearch});


\If{\textnormal{Eq.(\ref{eqTerminalCond}) is} satisfied}{
    $\mathbf{a}_{opt}=\mathbf{a}_{i+1}$ and \textbf{return} $\mathbf{a}_{opt}$; 
}

$i=i+1$;

\tcc{\small Acceleration by Broyden update}

Evaluate the Jacobian $\mathbf{J}_{i-1}$ (i.e., Eq.(\ref{eqJacobianDiff}));

Determine the inverse $\mathbf{J}_{i-1}^{-1}$ by SVD;

\While{$i<i_{max}$}{

Update $\mathbf{J}_{i}^{-1}$ by Eq.(\ref{eqInverseJacobianUpdate});

Compute $\mathbf{a}_{i+1}$ by Eq.(\ref{eqBroydenIteration});

\If{$D(\mathcal{T},\mathcal{S}(\mathbf{a}_{i+1})) > D(\mathcal{T},\mathcal{S}(\mathbf{a}_i))$}{

\textbf{break};

}

$\mathbf{a}_{opt}=\mathbf{a}_{i+1}$ and $i=i+1$;

}

}

\textbf{return} $\mathbf{a}_{opt}$;

\end{algorithm}


\begin{figure}[t] 
\centering
\includegraphics[width=.75\linewidth]{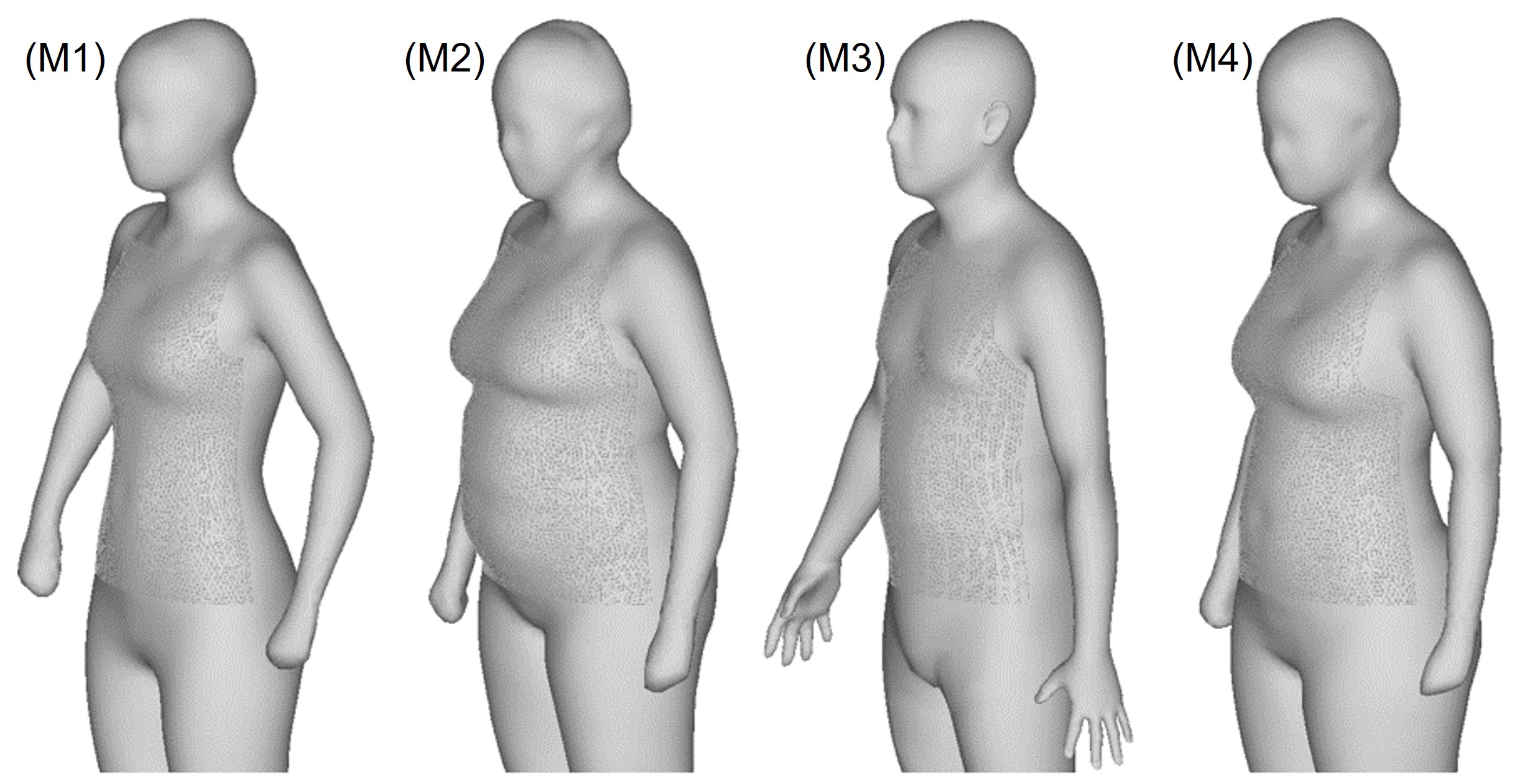}\\
\vspace{-5pt}
\caption{Four target models employed in our experimental tests.}\label{fig:TargetModels} 
\end{figure} 

\section{Results}\label{secResult}
The effectiveness of our algorithm has been verified on both the hardware setup (Section~\ref{subsecHardware}) and a simulation based platform by using the Abaqus FEA software. All the experimental tests were conducted on a laptop PC with a 2.6GHz CPU (Intel i7-9750H) and 32GB RAM. Four different human bodies obtained from a data-driven parametric human modeling approach\footnote{By online generators: \url{https://zishun.github.io/projects/3DHBGen/} (female models) and \url{http://humanshape.org/AdultShape/standingmale/} (male models).}~\cite{Chu2010,Reed2014} were employed as target models in our tests (see Fig.\ref{fig:TargetModels}). 

\subsection{Verification by Simulation}\label{subsecResultAlg}
We have developed a simulation based virtual platform to conduct the algorithm verification so that the influence from environmental factors (such as frictions, material hysteresis, camera instability, etc.) can be minimized. Two Python programs were developed to realize this virtual platform\rev{ (as illustrated in Fig.\ref{fig:DigitalPlatformWorkFlow})}{}, where one was the deformation control algorithm and the other was a plug-in for Abaqus generating deformed mesh surfaces that were converted into a point cloud for the virtual scanning result. The Ogden hyper-elastic model was employed in our simulation by choosing the silicone material -- Dragon skin: the shear modulus $G  = 75.449~\mathrm{KPa}$, the strain hardening exponent $\alpha = 5.836$ and the Poisson's ratio $\nu=0.4999$.

\begin{table}[t]\footnotesize
\centering
\caption{Computational Costs on Virtual Platform}
\begin{tabular}{c|c|c|c|c}
\hline\hline
\textbf{Models} & $M1$ & $M2$ & $M3$ & $M4$ \\ 
\hline 
\hline 
\multicolumn{5}{c}{Gradient-Descent Based Solver}\\
\hline 
\# of Steps & $30$ & $25$ & $14$ & $21$ \\ 
\hline 
Total Time (min.) & $2,310$ & $1,925$ & $1,078$ & $1,617$ \\ 
\hline 
\hline 
\multicolumn{5}{c}{Hybrid Solver with Accelerated Update}\\
\hline 
\# of Steps & $9$ & $17$ & $8$ & $12$ \\ 
Broyden Update$^\dag$ & $3,4,5$ & $2,3,4,6,7$ & $2,3,4,6$ & $2,3,4$ \\
\hline 
Total Time (min.) & $495$ & $979$ & $352$ & $726$ \\ 
\hline\hline
\end{tabular}
\begin{flushleft}
$^\dag$The steps that Brodyen updates instead of gradient-descent is applied. Note that $\tau=1.0\%$ and $i_{\max}=50$ are used in this experiment. 
\end{flushleft}\label{tab:CompStatistics}
\end{table}

\begin{figure}[t] 
\centering
\includegraphics[width=\linewidth]{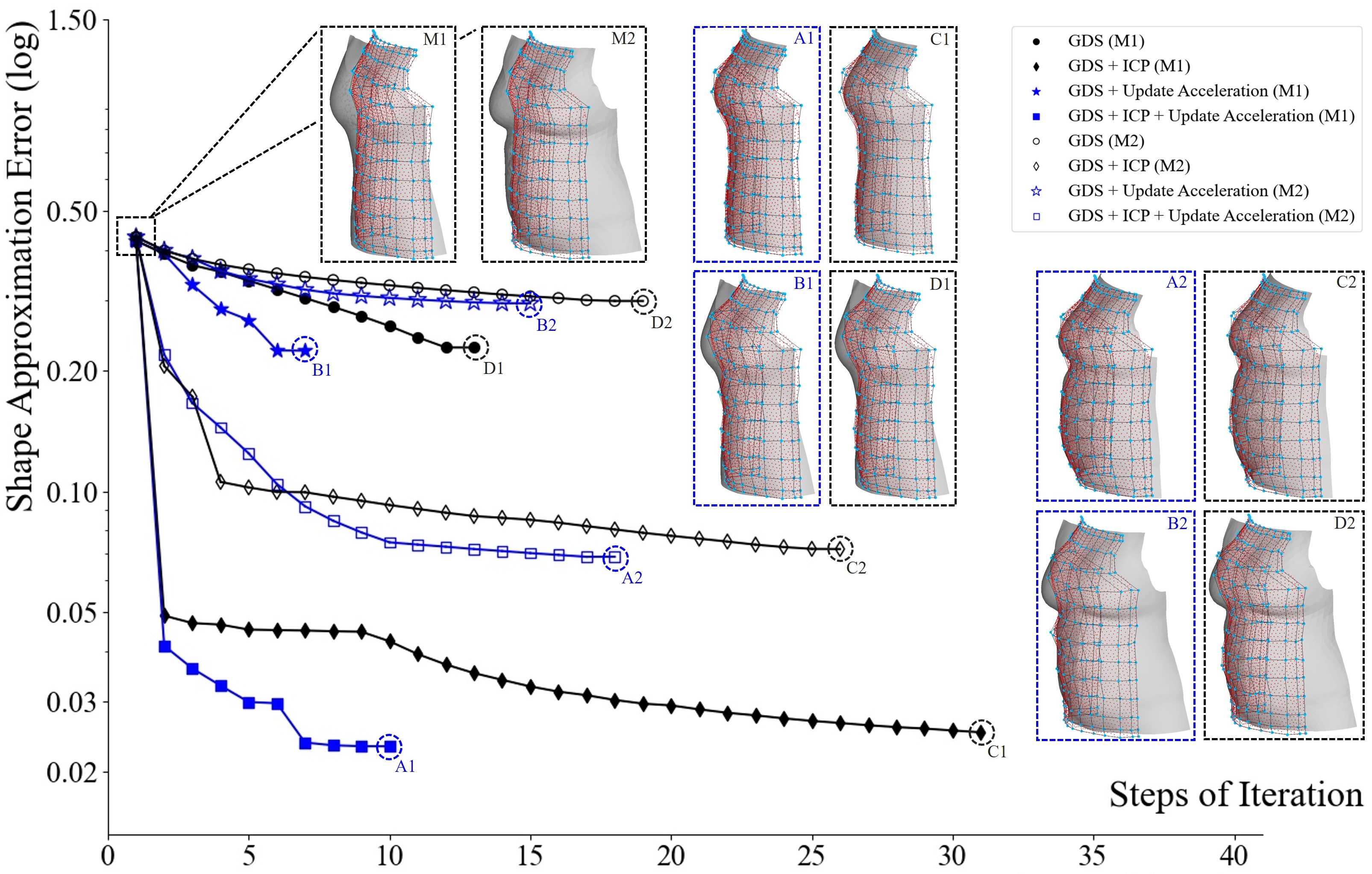}\\ 
\vspace{-5pt}
\caption{The ablation study taken on $M1$ and $M2$\rev{}{~for enhanced gradient-descent solver (GDS)}. Note that we use $\tau=1.0\%$ and $i_{\max}=50$ in this experiment to generate more steps of iteration. 
}\label{fig:AblationStudyResult} 
\end{figure}

The first experiment was for an ablation study to demonstrate the functionality of the ICP-based pose estimation and the acceleration based on the Broyden update. Different combinations of our algorithm's modules 1) with vs. without post estimation and 2) with vs. without update acceleration were tested on M1 and M2 models. From the curves shown in Fig.\ref{fig:AblationStudyResult}, we observed that the iteration without pose estimation \rev{always stoped}{stopped} at a relatively high level of shape approximate error. It was also interesting to find the hybrid solver with acceleration update converged with fewer steps of iteration. This was mainly \rev{caused for the reason that}{due to} the Broyden update\rev{}{ which} can actually adjust the optimizer to move in a direction that converges more quickly. The real computing time of the hybrid solver was even faster because much fewer simulation steps were needed to complete the evaluation of the numerical difference -- in short, 
$4.67 \times$ and $1.97 \times$ acceleration were achieved on $M1$ and $M2$ respectively.
%

Computational statistics for all models are given in Table \ref{tab:CompStatistics}. The steps applied for the Broyden update in comparison to the gradient-descent are reported. It was observed that these acceleration steps usually occurred in the early phase of the iterations (i.e., the first 10 steps). In other words, this acceleration technique was very effective so the shape approximation errors were reduced rapidly in the first few steps of the iterations.

\begin{table}[t]\footnotesize
\centering
\caption{Time Costs on Hardware Platform}
\begin{tabular}{c|c|c|c|c}
\hline\hline
\textbf{Models} & $M1$ & $M2$ & $M3$ & $M4$ \\ 
\hline 
\hline 
\# of Steps$^\dag$ & $7$ & $7$ & $10$ & $8$ \\ 
Broyden Update$^\ddag$ & $2,3,4$ & $2,3$ & $2,3,4,5,6,8,10$ & $2,3$ \\
\hline 
Total Time (min.) & $17.12$ & $20.30$ & $15.17$ & $27.08$ \\ 
\hline 
\end{tabular}
\begin{flushleft}
$^\dag$The hybrid solver is applied for deformation control on the hardware setup.\\
$^\ddag$The steps that Brodyen update instead of gradient-descent is applied.
\end{flushleft}\label{tab:TimeStatisticsHardware}
\end{table}

\begin{figure}[t] 
\centering
\includegraphics[width=\linewidth]{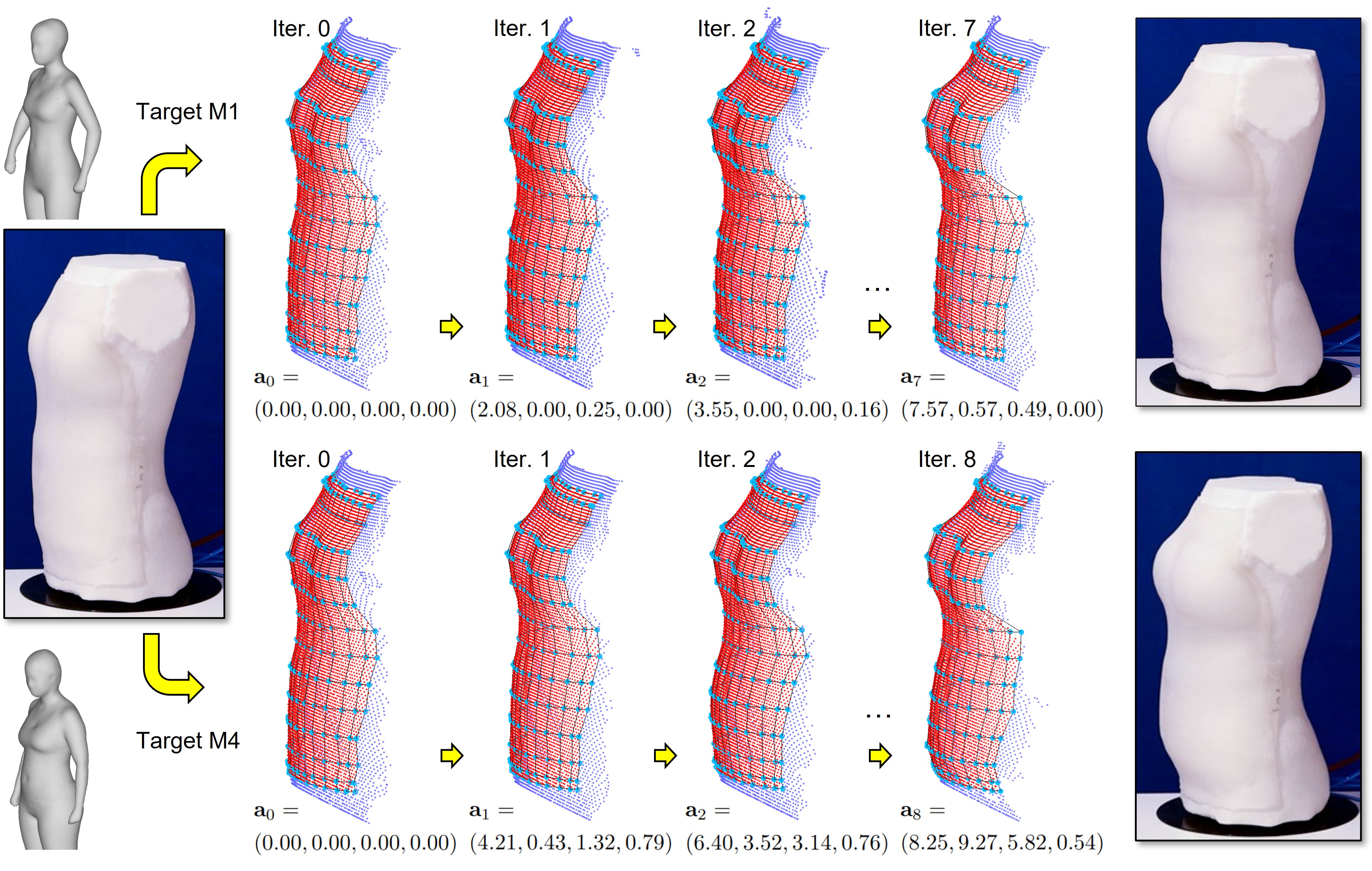}\\
\vspace{-8pt}
\caption{The progressive results of deformation with targets as $M1$ and $M4$\rev{}{, and the actuation values $\{\mathbf{a}_i\}$ (unit: $\mathrm{kPa}$) in different steps are given}.
}\label{fig:ProgressResults_M1M4} 
\end{figure}

\begin{figure}[t] 
\centering
\includegraphics[width=\linewidth]{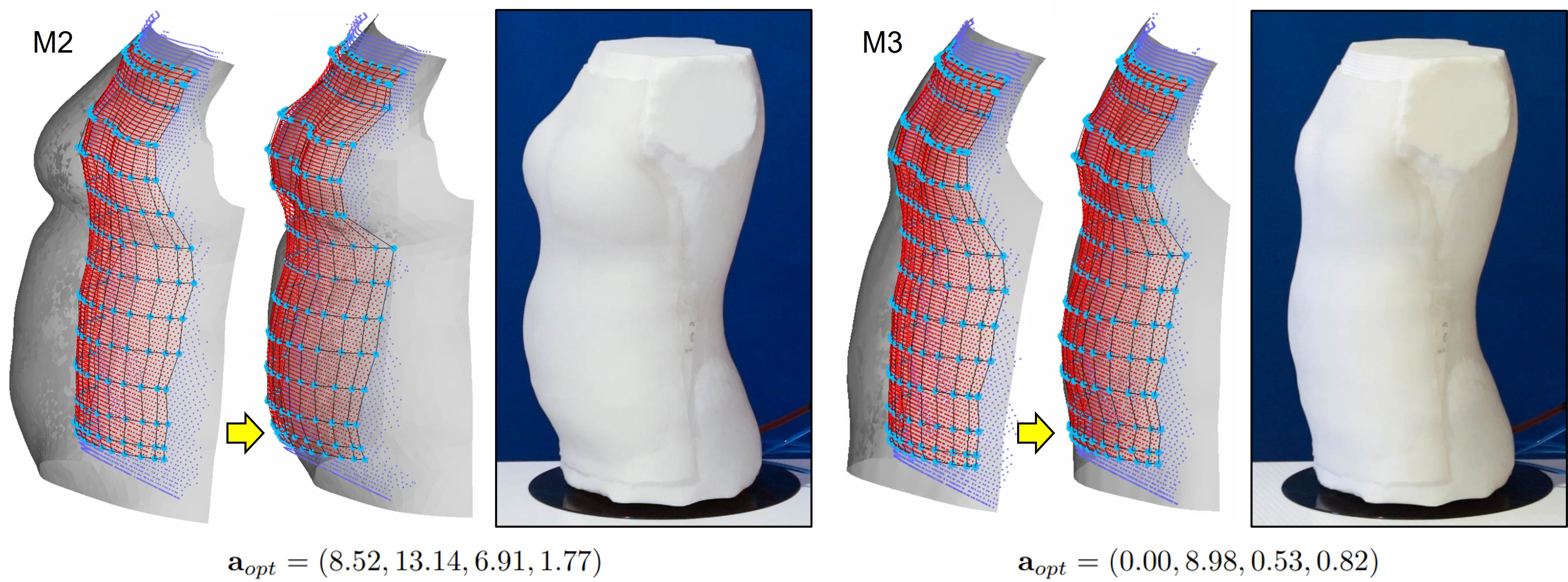}\\
\vspace{-8pt}
\caption{The experimental results by applying our deformation control algorithm on two examples $M2$ and $M3$, \rev{}{where} the target shapes (as gray surfaces) \rev{}{and the optimized actuation $\mathbf{a}_{opt}$ (unit: $\mathrm{kPa}$) are given}.
}\label{fig:FinalResults_M2M3} 
\end{figure}

\begin{figure}[!t]
\includegraphics[width=0.97\linewidth]{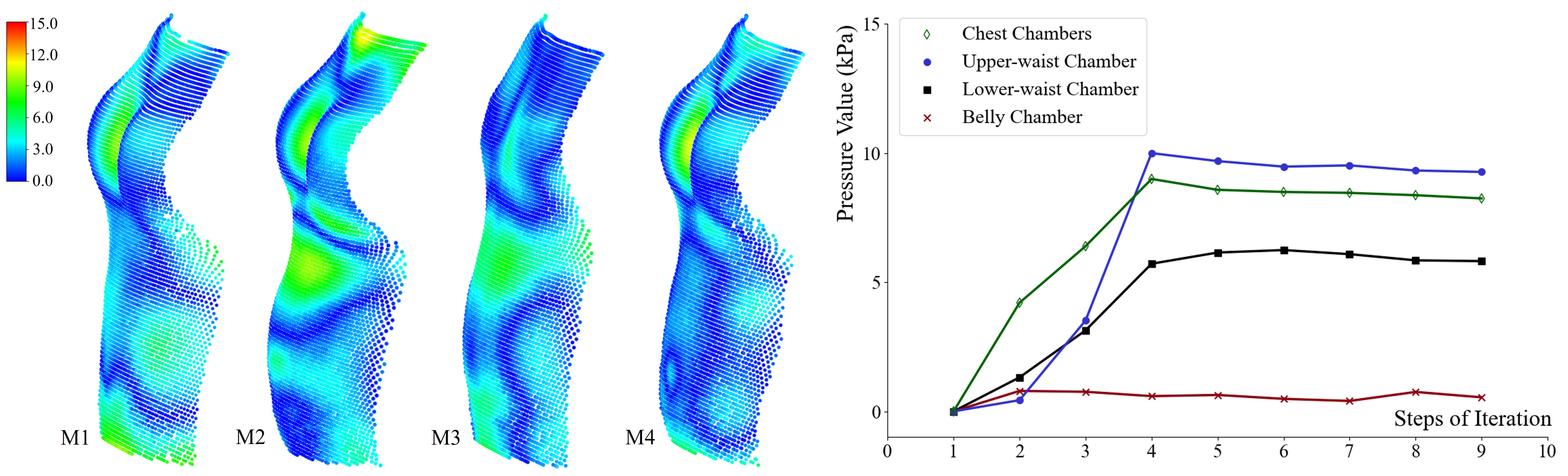}\\
\vspace{4pt}
\scriptsize 
\begin{tabular}{r|c|c|c|c}
\hline
\rev{}{\textbf{Models:}} & $M1$ & $M2$ & $M3$ & $M4$ \\ 
\hline\hline
\rev{}{\textbf{Average Err.:} $\frac{1}{n}\sum_{j=1}^n \| \mathbf{p}_j - \mathbf{c}^{\mathcal{T}}_j \|$} & $2.87$ & $3.28$ & $2.71$ & $2.31$ \\ 
\hline 
\rev{}{\textbf{Max. Err.:} $\max_{j=1, \ldots, n} \| \mathbf{p}_j - \mathbf{c}^{\mathcal{T}}_j \|$} & $9.98$ & $11.19$ & $8.25$ & $10.95$ \\ 
\hline 
\end{tabular}
\vspace{-5pt}
\caption{Shape approximation errors as color-maps (unit: mm) \rev{}{and statistics. Curves in the right give the change of pressures in different chambers during iteration for $M4$.}}\label{fig:FinalResults_ErrColorMap} 

\end{figure}

\subsection{Physical Experiment}\label{subsecResultAlg}
Our deformation control algorithm was implemented in C++ so that the real-time communication with hardware could be achieved. The tests were conducted on the examples with target shapes illustrated in Fig.\ref{fig:TargetModels}. Unlike the virtual platform based on FEA simulation, the total time to realize a target shape in hardware was much faster. This was because the computation conducted in Abaqus was much slower than the real actuation for achieving a stable shape on the hardware setup (i.e., $660.0$~sec. vs. $12.5$~sec. per actuation \rev{in}{on} average). The statistics of time cost on hardware setup are presented in Table \ref{tab:TimeStatisticsHardware}. 

We applied the proposed hybrid algorithm of deformation control to our hardware setup of the soft robotic mannequin. The results of our experimental tests were very encouraging. The computation could converge quickly in all examples. The progressive results obtained for the target models $M1$ and $M4$ during the iteration of deformation control are shown in Fig.\ref{fig:ProgressResults_M1M4}. The final results for the other two examples $M2$ and $M3$ are presented in Fig.\ref{fig:FinalResults_M2M3}. We also present the distribution of shape approximation errors on all four examples in Fig.\ref{fig:FinalResults_ErrColorMap}.

\begin{figure}
\centering
\includegraphics[width=\linewidth]{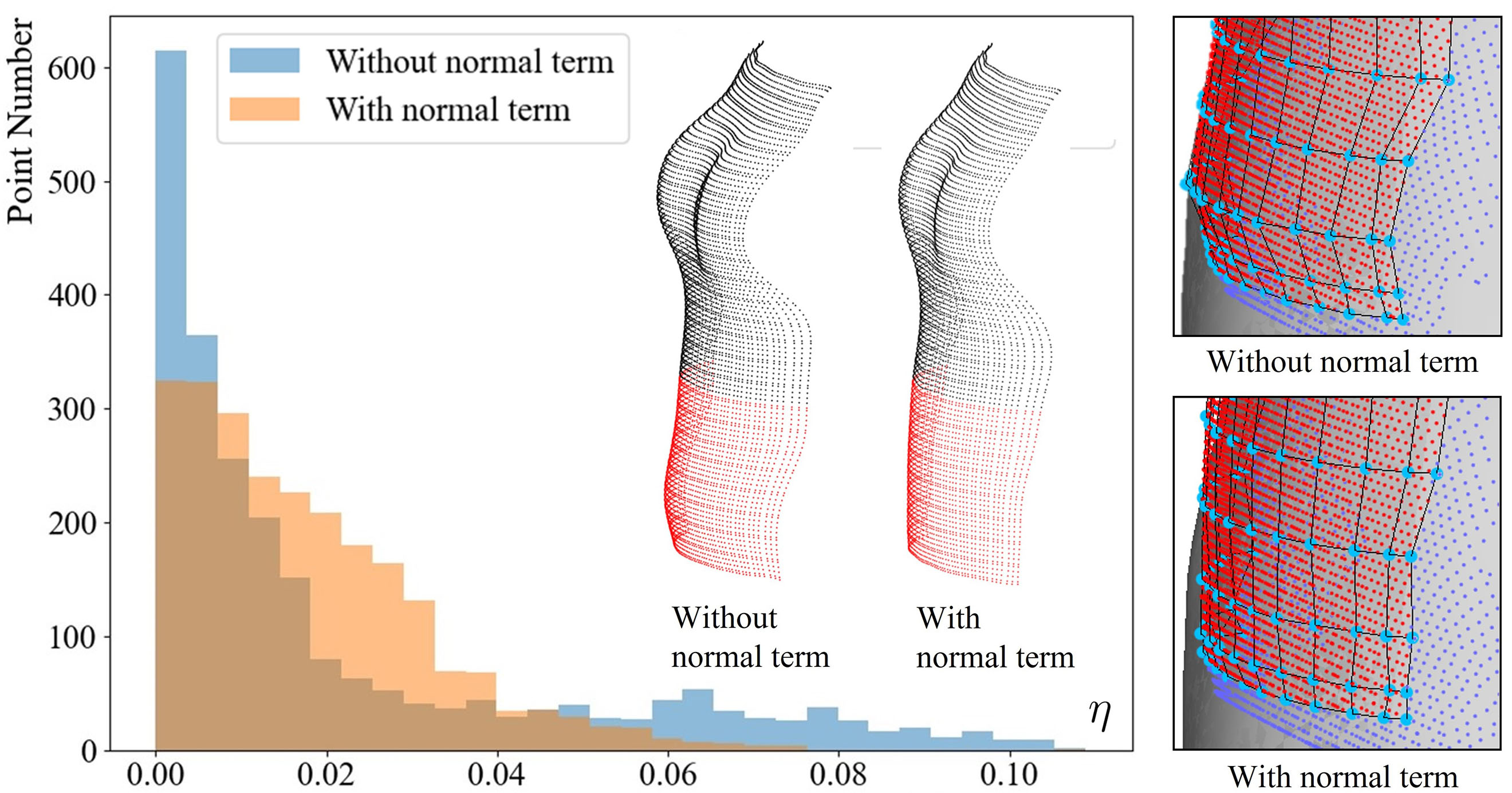}\\
\vspace{-5pt}
\caption{The study to show the effectiveness of the normal term in Eq.(\ref{eqObjFunc}), where the histogram of normal variation is evaluated at all sample points $\mathbf{p}_j$ in the waist region as $\eta$ (Eq.(\ref{eqNormalVarHistogram})).
}\label{fig:Results_EffectivenessNormTerm} 
\vspace{-5pt}
\end{figure}
\begin{figure}
\centering
\includegraphics[width=\linewidth]{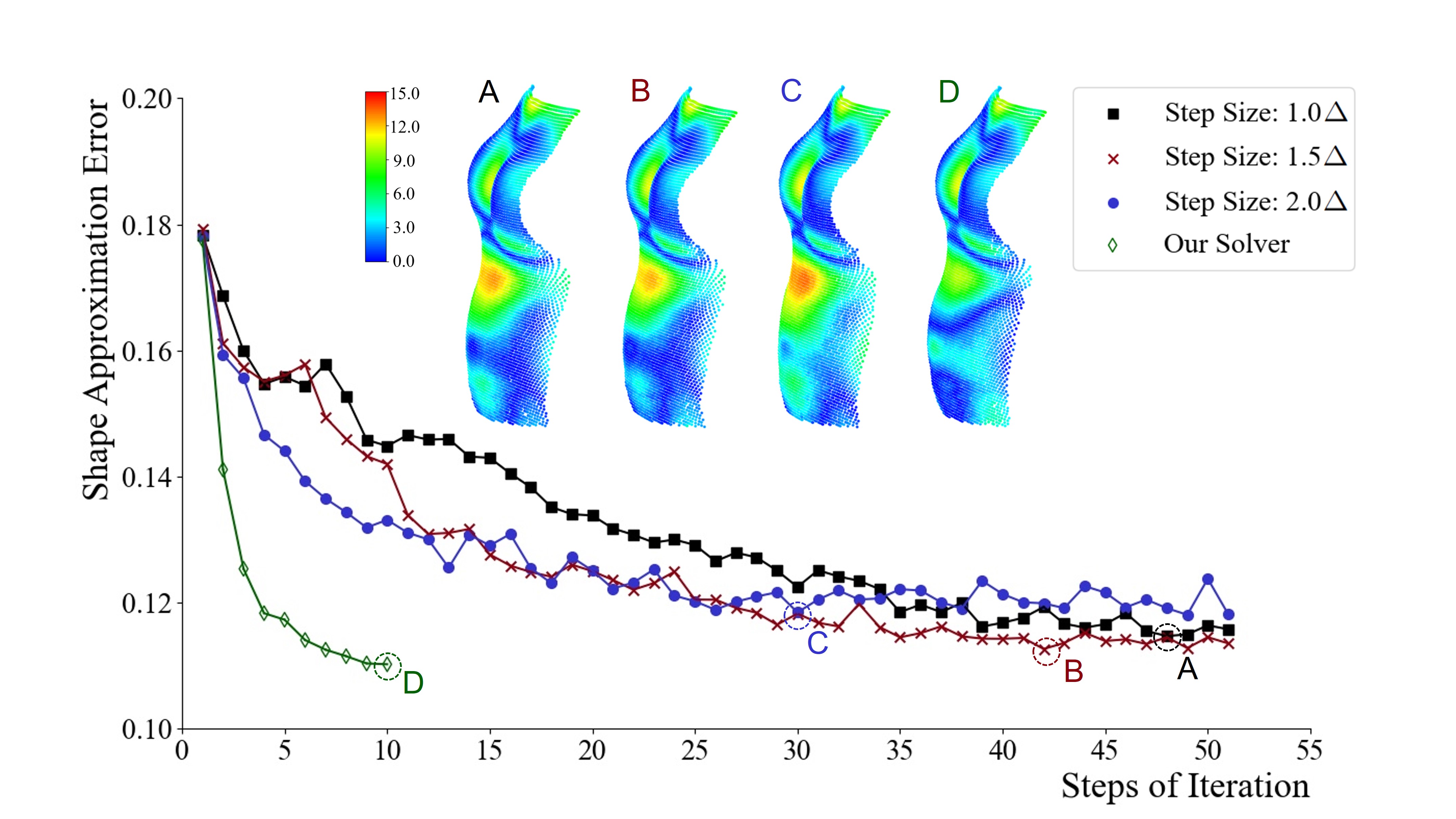}\\
\vspace{4pt}
\scriptsize 
\begin{tabular}{r|c|c|c|c}
\hline
\rev{}{\textbf{Different Results:}} & $1.0\Delta$ & $1.5\Delta$ & $2.0\Delta$ & Ours \\ 
\hline\hline
\rev{}{\textbf{Average Err.:} $\frac{1}{n}\sum_{j=1}^n \| \mathbf{p}_j - \mathbf{c}^{\mathcal{T}}_j \|$} & $3.84$ & $3.64$ & $4.42$ & $3.25$ \\ 
\hline 
\rev{}{\textbf{Max. Err.:} $\max_{j=1, \ldots, n} \| \mathbf{p}_j - \mathbf{c}^{\mathcal{T}}_j \|$} & $13.04$ & $12.65$ & $13.54$ & $11.27$ \\ 
\hline 
\end{tabular}
\caption{\rev{}{A conventional control algorithm akin to \cite{Okamura_CLoseLoopShapeControl} with different step sizes is tested to compare its results with ours on the $M2$ model, where $\Delta = 0.25$ -- the same value as our gradient evaluation (i.e., Eq.(\ref{eqGradientComponentf})) is adopted. Note that, our algorithm reaches the optimized result by 50 actuations in total.}
}\label{fig:Results_ControlAlgComparison} 
\vspace{-5pt}
\end{figure}

Another interesting study was regarding the significance of the normal term in Eq.(\ref{eqObjFunc}). When only the position term was applied, the iteration of our deformation control algorithm could become stuck at \rev{}{the} local optimum. The normal term can further enforce the shape information to the first order so that a higher similarity is obtained. The experimental results on $M1$ with vs. without the normal term are illustrated in Fig.\ref{fig:Results_EffectivenessNormTerm}. It was found that a `flat' lower waist region was formed by reinforcing the normal consistency between the physical mannequin (displayed as the point cloud) and the target model. A statistical analysis of the normal variation on sample points in the waist region was conducted and the results are displayed as the histogram of
\begin{equation}\label{eqNormalVarHistogram}
    \eta(\mathbf{p}_j) = 1-\mathbf{n}(\mathbf{p}_j) \cdot \mathbf{n}(\mathbf{c}^{\mathcal{T}}_j)
\end{equation}
with $\mathbf{c}^{\mathcal{T}}_j$ being the closest point of $\mathbf{p}_j$ on the target model $\mathcal{T}$. Note that the weight $\varpi = 3.0 \times 10^{-3}$ was employed for all our examples, which can well balance the importance of the position and the normal terms.

The advantage of our method was demonstrated by comparing to a conventional control strategy that incrementally changes the actuation parameters with a fixed step size \cite{Okamura_CLoseLoopShapeControl}. As shown in Fig.\ref{fig:Results_ControlAlgComparison}, the results for $M2$ generated by the incremental control algorithm with different step sizes had larger shape approximation errors than ours. This was mainly caused by the complex non-convex shape of the mannequin and the interference between overlapped chambers. Differently, our algorithm can generate better shape control in those non-convex regions.
In order to apply the conventional control strategy, the ICP-based pose estimation has been excluded from these tests, where we used a fixed and optimized pose for the template $M2$ model. Therefore, our result in Fig.\ref{fig:Results_ControlAlgComparison} is slightly different from Fig.\ref{fig:FinalResults_ErrColorMap}.

\begin{figure}[t] 
\centering
\includegraphics[width=\linewidth]{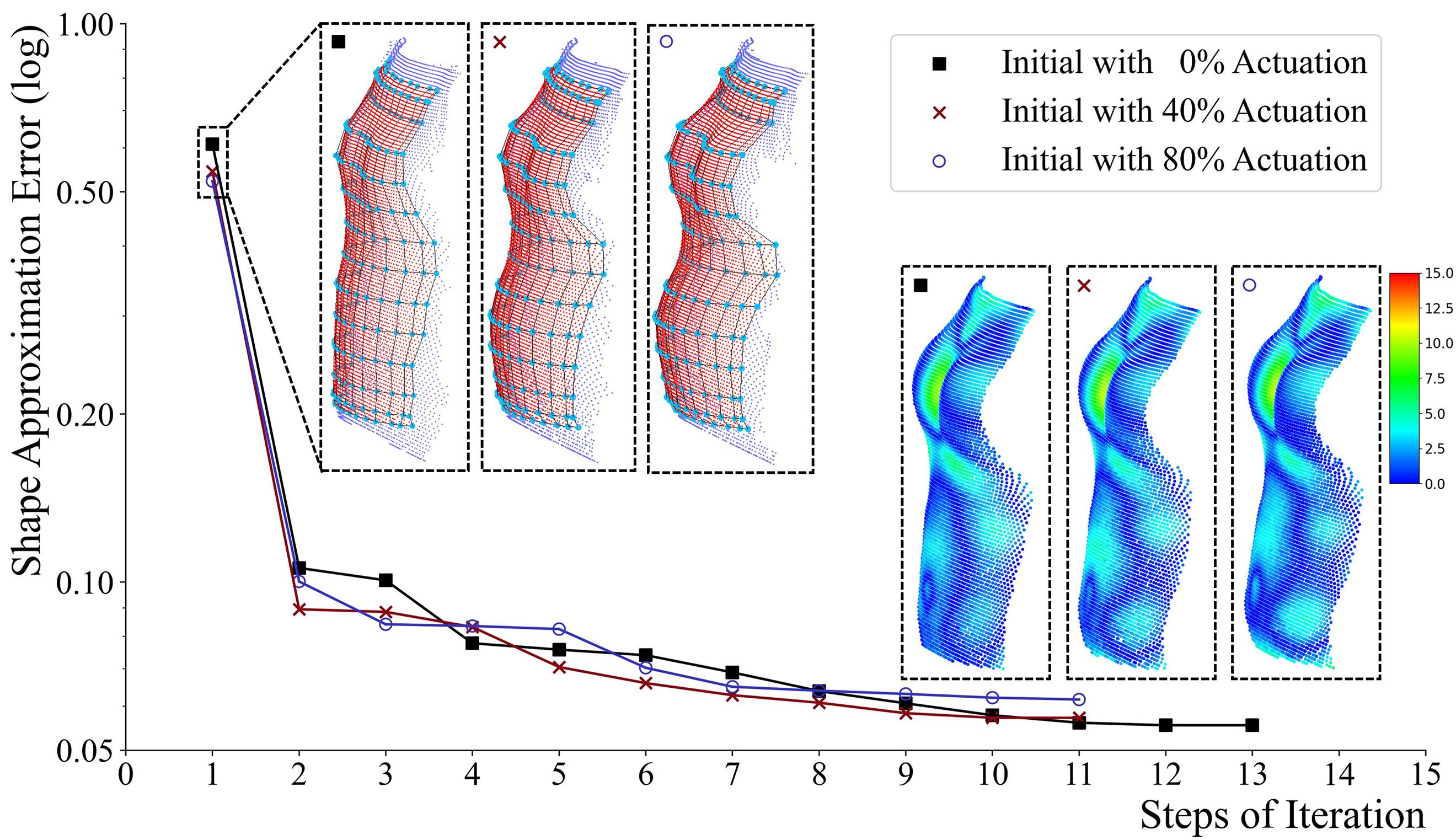}\\
\vspace{-5pt}
\caption{A study about using different initial guesses for the computation in deformation control -- similar but slightly different rates of convergence are observed, where shape approximation errors are visualized by color-maps. 
}\label{fig:Results_InitialGuessStudy} 
\vspace{-5pt}
\end{figure}

The last experiment was to study how the computation \rev{were}{was} affected by the initial guess. This experiment was conducted on $M4$ as it had large shape variation compared to the mannequin's shape before actuation. As shown in Fig.\ref{fig:Results_InitialGuessStudy}, we used three different initial shapes in our deformation control algorithm -- a slim one with zero actuation on all chambers and the other two with $40\%$ and $80\%$ full actuation on all chambers respectively. From the curves of computation, we observed \rev{an}{a} similar speed of convergence although \rev{it was in general}{it was, in general,} easier to converge in a shrinkage process from a `fat' into `slim' body shape rather than an inflating process -- i.e., the initial guess with zero actuation takes more steps to converge. 

\section{Conclusion and Discussion}
A novel soft robotic system as a deformable mannequin has been presented in this paper to physically realize the 3D shapes of different human bodies. The novelty of our system in comparison to existing deformable mannequin robotic systems is that we employ pneumatic actuation to deform soft chambers that change the soft membrane on the mannequin. We have presented our new design and corresponding fabrication methods. A new algorithm has been introduced to solve the challenging issue of deformation control. 
Vision feedback by \rev{}{a} 3D scanner has been employed in our system to evaluate the shape approximation error. 
Our algorithm was formulated under the framework of gradient-descent. By including the steps of pose estimation and Broyden-update acceleration, the algorithm has demonstrated fast convergence and can reduce the time-consuming steps of derivative estimation on the hardware. Physical experiments have been conducted to verify the efficiency and effectiveness of our algorithm.

One major limitation of our approach is the slow response speed of chambers caused by material hysteresis, which leads to time-consuming operations to evaluate numerical differences on the hardware setup. \rev{}{The responsiveness can also be improved by employing fast digital pressure regulators.} A further limitation is that the convergence rate is dependent on the initial estimates. This is a common problem of optimization-based methods. We plan to use a machine-learning strategy together with task space sampling to solve this problem in the future. Lastly, although the \rev{structure-light}{structured-light} based 3D scanner is accurate, there will be a greater challenge when vision obstacles (e.g., clothes) are presented. We plan to use the proprioception technique enabled by sensor-integrated chambers \cite{Scharff2021TMECH} in our future work.


{\small
\bibliographystyle{IEEEtran}
\bibliography{reference.bib}
}

\end{document}